\documentclass{article}

\usepackage{microtype}
\usepackage{graphicx}
\usepackage{subcaption}
\usepackage{booktabs}

\usepackage{hyperref}

\usepackage{multirow}
\usepackage{enumitem}

\usepackage[accepted]{icml2026_weightsymmetry}

\usepackage{amsmath}
\usepackage{amssymb}
\usepackage{mathtools}
\usepackage{amsthm}

\usepackage[capitalize,noabbrev]{cleveref}

\theoremstyle{plain}

\theoremstyle{definition}

\theoremstyle{remark}

\usepackage[textsize=tiny]{todonotes}

\begin{document}

\twocolumn[
  \icmltitle{Shortcuts in the Tail: Debiasing via Post-Hoc Spectral Compression of Fine-Tuning Updates}
  \begin{icmlauthorlist}
    \icmlauthor{Edward Sun}{ucla}
    \icmlauthor{Dmitrii Troitskii}{independent}
  \end{icmlauthorlist}
  \icmlaffiliation{ucla}{Department of Computer Science, UCLA, Los Angeles, CA, USA}
  \icmlaffiliation{independent}{Independent}
  \icmlcorrespondingauthor{Edward Sun}{edwardsun12895@g.ucla.edu}
  \icmlcorrespondingauthor{Dmitrii Troitskii}{troitskii.d@northeastern.edu}
  \icmlkeywords{Machine Learning, ICML}
  \vskip 0.3in
]

\printAffiliationsAndNotice{}

\begin{abstract}
Fine-tuning often introduces spurious correlations alongside task knowledge – models learn to rely on shortcuts, causing systematic failures. Existing mitigations require retraining, group labels, or curated counterfactual data. We show a simple post-hoc intervention to reduce shortcut reliance without modifying training or data setup: truncating the tail of the SVD of $\Delta W = W_\mathrm{ft} - W_\mathrm{base}$ decreases the spurious-group gap while preserving task accuracy. Across three instruction-tuned models and four classification benchmarks, top-$k$ truncation reduces the gap by up to $5\times$ on the CivilComments dataset at $<\!2$ pp accuracy loss. We propose this works because the shortcut response sits in the tail of the singular ordering of $\Delta W$, a claim about how truncation behaves rather than about the raw singular values, which are broadly distributed and look the same across all four datasets. A controlled boundary case in which fine-tuning has only a shortcut to learn shows the predicted FT-to-base collapse, and bottom-/random-$k$ and matched-rank LoRA controls rule out generic low-rank approximation and rank-constrained training as the explanation. We read this as preliminary evidence that the singular basis of $\Delta W$ is a useful coordinate system for studying what fine-tuning has learned.
\end{abstract}

\section{Introduction}

\begin{figure*}[ht!]
    \centering
    \includegraphics[width=0.8\textwidth]{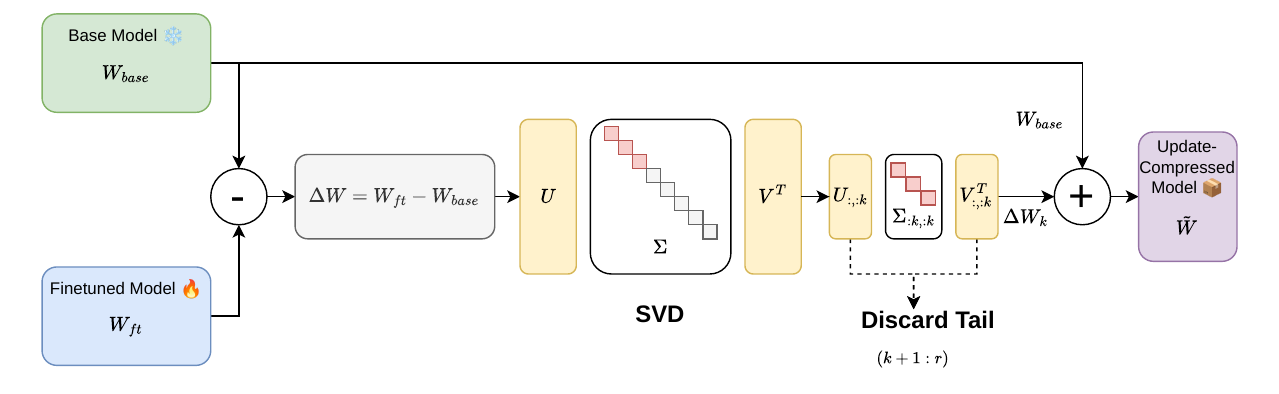}
    \caption{\textbf{Post-hoc spectral compression of fine-tuning updates.} For each weight matrix, compute $\Delta W = W_\mathrm{ft} - W_\mathrm{base}$, take its SVD $\Delta W = U \Sigma V^\top$, keep only the top $k$ singular values, and reconstruct $\widetilde{W} = W_\mathrm{base} + U_{:,:k} \Sigma_{:k,:k} V_{:,:k}^\top$. No retraining, data, or group labels; debiasing comes from \emph{which} singular directions are kept.}
    \label{fig:wide_figure}
\end{figure*}

Fine-tuning instruction-tuned LLMs often achieves high in-distribution accuracy by exploiting spurious correlations \citep{civilcomments, civilcommentswilds, HANS, PAWS}, causing systematic failure on underrepresented groups and adversarial inputs \citep{wu2022generatingdatamitigatespurious, varma2024ravldiscoveringmitigatingspurious, zhou2024explorespuriouscorrelationsconcept, yang2025escapingspuriverselargevisionlanguage, chen2026safetymiragespuriouscorrelations, wang2025biaspretendstruthspurious, salles2025lorausersbewarespurious, shuieh2025assessingrobustnessspuriouscorrelations}. Existing mitigations intervene during training or on the data itself \citep{sagawa2020distributionallyrobustneuralnetworks, wu2022generatingdatamitigatespurious, chen2026safetymiragespuriouscorrelations, zou2025representationengineeringtopdownapproach}: retraining with a reweighted loss that upweights minority groups, augmenting the training set with synthetic counterfactual examples, or modifying intermediate representations during training. All require either group labels, curated counterfactual data, or a full retraining loop, and none directly examines how the shortcut is stored. We ask a structural question instead: does the fine-tuning update itself encode the distinction between task signal and shortcut?

We analyze the difference $\Delta W = W_\mathrm{ft} - W_\mathrm{base}$ between the fine-tuned and base weights, decomposing it into its singular value decomposition $\Delta W = U \Sigma V^\top$ across three instruction-tuned models. We find that \textbf{truncating the tail of this decomposition selectively removes shortcut reliance while preserving task accuracy}. The claim is about the singular basis as an ordered coordinate system: truncation behaves as if task-relevant and shortcut-relevant directions occupy different parts of the ordering, even though the raw singular values $\sigma_i$ are broadly distributed and show no visible separation. The structure is recovered from the effect of intervention rather than read off the spectrum directly.

This yields a label-free, retraining-free debiasing method together with a sharp prediction. Unlike prior SVD work targeting base-model efficiency \citep{wang2025svdllmtruncationawaresingularvalue, wang2025svdllmv2optimizingsingular, hsu2022languagemodelcompressionweighted}, low-rank training \citep{hu2021loralowrankadaptationlarge}, or task-arithmetic analyses \citep{jain2024mechanisticallyanalyzingeffectsfinetuning, ilharco2023editingmodelstaskarithmetic}, we compress the update post-hoc to target the tail. Decoupling appears across all four natural-shortcut datasets as a matter of degree: sharpest on CivilComments (up to $5\times$ gap reduction at $<\!2$ pp accuracy loss), visible but more modest on MNLI, FEVER, QQP. The hypothesis predicts a sharp boundary: if fine-tuning has no signal except the shortcut, no top-vs-tail structure exists and the only debiasing route is to collapse $\Delta W$ toward an unbiased base. A controlled IMDB-marker setting realises this regime (Sec.~\ref{sec:results:mechanism}). Bottom-$k$, random-$k$, and matched-rank LoRA controls rule out generic low-rank approximation and rank-constrained training.

\paragraph{Contributions.} (1) A label-free, retraining-free debiasing method based on post-hoc top-$k$ SVD of $\Delta W$, reducing the gap on every (model, dataset) cell at $<\!2$ pp accuracy loss, by up to $5\times$ on CivilComments. (2) A behavioural mechanism (shortcut response in the tail of the singular ordering), with decoupling visible across all four natural-shortcut datasets and sharpest on CivilComments. (3) A controlled IMDB setting realising the predicted boundary: a bidirectionally perfect injected marker is the only signal SFT can learn, so $\Delta W$ encodes the shortcut alone. With no top-vs-tail structure to exploit, top-$k$ can only shrink $\Delta W$ toward zero, returning the model to its (unbiased, accurate) base; gap and accuracy therefore lockstep along an FT-to-base trajectory. Bottom-/random-$k$ and matched-rank LoRA rule out generic low-rank approximation and rank-constrained training.

\section{Method}
\label{sec:method}

\paragraph{Models and tasks.} We evaluate Qwen2.5-0.5B-Instruct, Gemma-3-1B-IT, and Qwen2.5-7B-Instruct on five classification tasks. CivilComments-WILDS \citep{civilcommentswilds} contains identity-group mentions co-occurring with toxic labels. MNLI \citep{mnli} is a natural language inference task where premise and hypothesis often share lexical content in entailment pairs, giving lexical overlap as a shortcut for predicting entailment. QQP \citep{qqp} is a paraphrase identification task where the two questions in a paraphrase pair tend to share high word overlap, again offering a lexical shortcut. FEVER \citep{fever} is a fact-verification task where claims and retrieved evidence often share large spans of text in supported claims, giving evidence-overlap as a shortcut. We use each dataset as-is, without filtering or rebalancing. An IMDB sentiment dataset with an injected prefix marker bidirectionally perfectly predictive of the negative class (present iff negative) serves as the boundary case: the marker is the only available signal, so SFT encodes nothing else. $\Delta W$ has no top-vs-tail structure to exploit, and post-hoc compression's only route is to shrink $\Delta W$ toward zero. The base model never saw the marker and is both unbiased and accurate on this distribution, so collapsing the update returns the model to a high-accuracy, low-gap point. All tasks use full-parameter SFT with three seeds. Evaluation uses group-balanced validation sets, reporting accuracy and the spurious-group gap $\Delta_\mathrm{gap} = \mathrm{Acc}_\mathrm{maj} - \mathrm{Acc}_\mathrm{min}$ ($\Delta_\mathrm{gap} \approx 0$ for an unbiased model).

\paragraph{Post-hoc compression.} For every 2D weight matrix (excluding biases and layer norms), let $\Delta W = U \Sigma V^\top$. At retention $\rho \in (0,1]$ we keep $k = \lceil \rho r \rceil$ singular values and reconstruct $\widetilde{W} = W_\mathrm{base} + U_{:,:k} \Sigma_{:k,:k} V_{:,:k}^\top$, evaluating without further training.

\paragraph{Controls.} \textbf{Bottom-$k$} keeps the smallest $k$ values; \textbf{random-$k$} selects $k$ uniformly at random. Together they isolate magnitude ordering from low-rank approximation.

\paragraph{LoRA comparison.} A LoRA \citep{hu2021loralowrankadaptationlarge} rank sweep on CivilComments at $r \in \{16, 32, 64, 128, 256\}$, $\alpha = 2r$, three seeds. The comparison tests \emph{post-hoc} truncation (unconstrained FT then drop the tail) against \emph{rank-constrained training} (optimizer packs task and shortcut into a fixed subspace from the start). We do not claim LoRA is a worse FT method, only that the spectral-tail structure post-hoc truncation exploits is absent in LoRA updates at matched rank.

\section{Results}
\label{sec:results}

We report \emph{bias reduction (\%)} $= 100(\Delta_\mathrm{ft} - \Delta_r)/|\Delta_\mathrm{ft}|$ and \emph{accuracy loss (pp)} $= 100(\mathrm{acc}_\mathrm{ft} - \mathrm{acc}_r)$. Trajectory plots are \emph{parametric in retention $r$}: each point is one $r$ as it sweeps $90\%\!\to\!5\%$, and neither axis is monotone in $r$. As $r$ decreases, trajectories first move \emph{up} (tail-truncation: bias drops at preserved accuracy), then \emph{right} (top components removed, accuracy collapses, model reverts toward base); the non-monotonicity reflects this regime transition, not noise.

\subsection{A sweet spot exists in every (model, dataset) cell}
\label{sec:results:sweet}

\begin{figure}[t]
  \centering
  \includegraphics[width=\columnwidth]{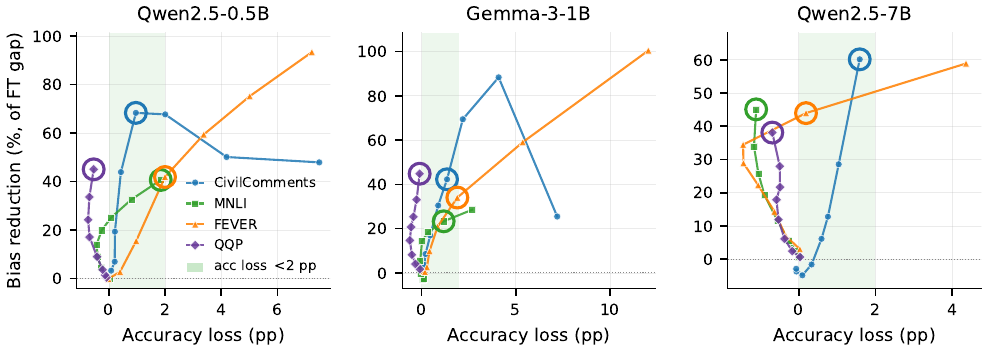}
  \caption{\textbf{Bias-vs-accuracy trajectories, parametric in retention $r$.} One panel per model. Each curve traces (accuracy loss, bias reduction) for one of CivilComments / MNLI / QQP / FEVER as $r$ sweeps $90\%\!\to\!5\%$. Green band: no-cost zone (accuracy loss $<\!2$ pp); hollow rings mark each dataset's sweet spot. The region to the left of the green band, where accuracy loss is negative, is also notable: as the model reverts toward an unbiased base, accuracy on the group-balanced evaluation can rise above the fine-tuned level, since the shortcut was hurting balanced accuracy in the first place. Curves move \emph{up} (tail-truncation: bias drops at preserved accuracy), then \emph{right} (top-truncation: accuracy collapses, model reverts toward base); the apparent non-monotonicity reflects this regime transition, not noise. Values exceeding $100\%$ at small $r$ indicate the residual gap has flipped sign as the model reverts to base, not super-debiasing; sweet spots are always $\le\!100\%$.}
  \label{fig:sweet}
\end{figure}

Table~\ref{tab:consistency} reports per-cell sweet-spot bias reduction: the maximum reduction inside the no-cost zone (accuracy loss $<\!2$ pp), with $r^*$ in parentheses. Top-$k$ reduces the gap on all 12 cells, from $23\%$ (MNLI/\textsc{Gemma-1B}) to $68\%$ (CivilComments/\textsc{Qwen-0.5B}); 11/12 exceed $30\%$. Across cells, $r^*$ ranges from $5\%$ to $20\%$, with \textsc{Qwen-7B} consistently benefiting from more aggressive truncation than the smaller models. Fig.~\ref{fig:sweet} traces full trajectories. Some pass $100\%$ at $r{=}5\%$ because the model has reverted close to base and the residual gap flips sign (over-correction, not super-debiasing), so we report the in-zone maximum.

\begin{table}[t]
  \centering
  \footnotesize
  \caption{\textbf{Empirical bias reduction at sweet-spot retention $r^*$} (max reduction with accuracy loss $<\!2$ pp; $r^*$ in parentheses). Direct measurements, not a mechanism decomposition (Sec.~\ref{sec:results:mechanism}). IMDB-marker excluded: in its boundary regime, accuracy moves sharply outside the no-cost zone (upward, toward base); see App.~\ref{app:imdb-control}.}
  \label{tab:consistency}
  \setlength{\tabcolsep}{3pt}
  \begin{tabular}{lcccc}
    \toprule
    Model & Civil & MNLI & FEVER & QQP \\
    \midrule
    \textsc{Qwen2.5-0.5B} & $+68$ {\scriptsize ($20\%$)} & $+41$ {\scriptsize ($5\%$)} & $+42$ {\scriptsize ($20\%$)} & $+45$ {\scriptsize ($5\%$)} \\
    \textsc{Gemma-3-1B}   & $+42$ {\scriptsize ($20\%$)} & $+23$ {\scriptsize ($10\%$)} & $+34$ {\scriptsize ($15\%$)} & $+45$ {\scriptsize ($5\%$)} \\
    \textsc{Qwen2.5-7B}   & $+60$ {\scriptsize ($5\%$)}  & $+45$ {\scriptsize ($5\%$)}  & $+44$ {\scriptsize ($10\%$)} & $+38$ {\scriptsize ($5\%$)} \\
    \bottomrule
  \end{tabular}
\end{table}

\subsection{Mechanism: ordering in the singular basis}
\label{sec:results:mechanism}

We propose a mechanism for the empirical result above and use IMDB-marker to expose its predicted boundary. Write $\Delta W = \sum_i \sigma_i u_i v_i^\top$. Top-$k$ truncation preserves the model's response in directions $v_1, \ldots, v_k$ and removes it in $v_{k+1}, \ldots, v_n$. That truncation can preserve accuracy while reducing the gap suggests a directional, behavioural claim: task-relevant inputs are predominantly served in the top right-singular vectors of $\Delta W$, shortcut-related inputs in the bottom. We call this the \emph{spectral-stratification hypothesis}, explicit that it is a claim about \emph{ordering} in the singular basis, recovered from the effect of truncation, not about energy concentration. The raw spectrum is broadly distributed yet truncation cleanly removes the bias-correlated component on CivilComments. We do not assert ``shortcuts have small singular values'', only that, behaviourally, truncating the smaller-$\sigma_i$ part preferentially removes shortcut reliance.

\begin{figure}[t]
  \centering
  \includegraphics[width=\columnwidth]{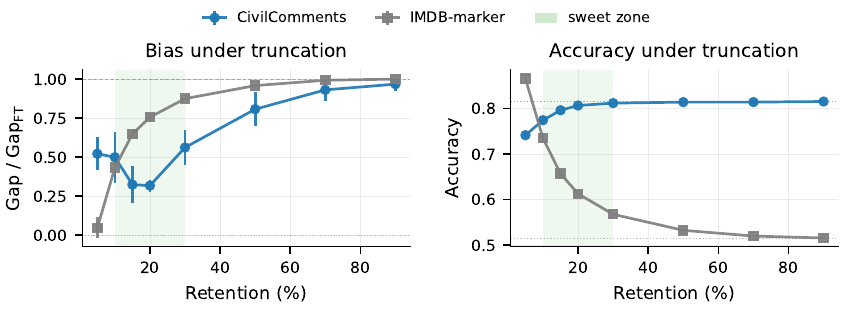}
  \caption{\textbf{Trajectory shape distinguishes the spectral picture from its boundary.} Normalized gap (left) and accuracy (right) vs.\ retention $r$, \textsc{Qwen-0.5B}. \textbf{CivilComments}: gap and accuracy \emph{decouple}. The gap drops through the sweet zone (green) to $\sim\!0.3\,\Delta_\mathrm{ft}$ while accuracy stays flat at FT level ($\sim\!0.81$). Spectral stratification predicts this: shortcut and task responses live in different parts of the singular basis, so removing the tail reduces one without disturbing the other. The same decoupling appears on MNLI, FEVER, QQP at smaller magnitude (Fig.~\ref{fig:sweeps-app}). \textbf{IMDB-marker}: gap and accuracy \emph{lockstep} along an FT-to-base trajectory. Accuracy \emph{rises} ($\sim\!0.51\!\to\!0.87$) while the gap \emph{falls} ($\Delta_\mathrm{ft}\!\to\!0$), meeting at the unbiased base. With the marker the only signal SFT can learn, no top-vs-tail structure exists; the only debiasing route is to collapse $\Delta W$ entirely. The two trajectory shapes (decoupling on natural-shortcut datasets, lockstep on IMDB-marker) are the diagnostic.}
  \label{fig:mechanism}
\vspace{-0.2in}
\end{figure}

\paragraph{IMDB-marker as a predicted boundary.} The hypothesis predicts a sharp boundary: when SFT has no signal except the shortcut, the entire update encodes it and no top-vs-tail structure exists for truncation to exploit. The only path top-$k$ can take is to shrink $\Delta W$ toward zero, returning the model to base ($\widetilde{W}=W_\mathrm{base}$ at $r=0$); the base model never saw the marker and is unbiased on this distribution. Both metrics improve in lockstep along an FT-to-base trajectory: accuracy \emph{rises} from $\sim\!0.51$ (FT, dominated by shortcut) toward $\sim\!0.87$ (base) and the gap \emph{drops} from $\Delta_\mathrm{ft}$ toward $\sim\!0$, meeting at the unbiased base. This is exactly what IMDB-marker shows: not a competing mechanism but the framework correctly identifying its own boundary, where selective debiasing reduces to global collapse of the update.

\paragraph{Empirical claim vs.\ mechanistic claim.} The empirical result of Table~\ref{tab:consistency} is direct measurement: top-$k$ truncation reduces the gap on every cell at $<\!2$ pp accuracy loss, independent of the spectral-stratification hypothesis. The mechanism claim is separate: we propose that truncation works because the shortcut response sits in the tail of the singular ordering of $\Delta W$. The predicted decoupling signature is visible across all four natural-shortcut datasets in Fig.~\ref{fig:sweeps-app}: every panel shows a flat blue accuracy curve at FT level while the red gap curve lifts toward base. The effect is sharpest on CivilComments and more modest on MNLI, FEVER, QQP, but trajectory shape is qualitatively the same. IMDB-marker realises the predicted boundary: with no task signal in the top components, $\Delta W$ has no separable structure and truncation can only return the model to base, so gap and accuracy lockstep rather than decouple. We do not claim a mechanism decomposition per cell: each natural-shortcut cell's reduction may reflect mostly selective tail removal, mostly partial reversion toward base, or a mixture, with relative weights likely varying across (model, dataset). Resolving this requires direct probes of the singular subspaces of $\Delta W$, left to future work.

\subsection{Comparison to alternatives and scaling}
\label{sec:results:methods}

\begin{figure}[t]
  \centering
  \includegraphics[width=\columnwidth]{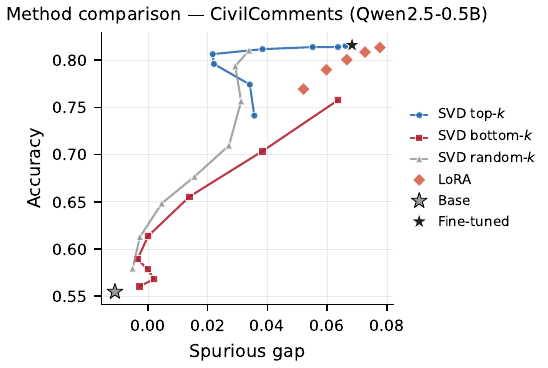}
  \caption{\textbf{Top-$k$ uniquely separates accuracy from bias; alternatives don't.} Accuracy versus gap on CivilComments (Qwen-0.5B), parametric in $r$ ($90\%\!\to\!5\%$). \textbf{Top-$k$ SVD} sweeps along the high-accuracy edge, reaching low gap before accuracy degrades. \textbf{Bottom-$k$} removes top components and accuracy collapses faster than the gap shrinks. \textbf{Random-$k$} sits between, ruling out generic low-rank approximation: magnitude ordering is what matters. \textbf{LoRA} at matched rank ($r\!\in\!\{16,32,64,128,256\}$) clusters near FT regardless of rank, never reaching the low-gap region.}
  \label{fig:methods}
  \vspace{-0.2in}
\end{figure}

Fig.~\ref{fig:methods} compares top-$k$ against three baselines on CivilComments. Bottom-$k$ removes top components and accuracy collapses faster than the gap; random-$k$ sits between. Together they rule out generic low-rank approximation: magnitude ordering matters, not dimension count. LoRA at matched rank does not reproduce post-hoc debiasing, clustering near FT regardless of rank. The spectral-tail structure post-hoc truncation exploits is a property of \emph{unconstrained} FT, where the optimizer can place dominant task patterns in a high-magnitude top subspace and let weaker shortcuts settle in the tail; with rank fixed in advance, the optimizer must pack both into the same subspace.

\begin{figure}[t]
  \centering
  \includegraphics[width=\columnwidth]{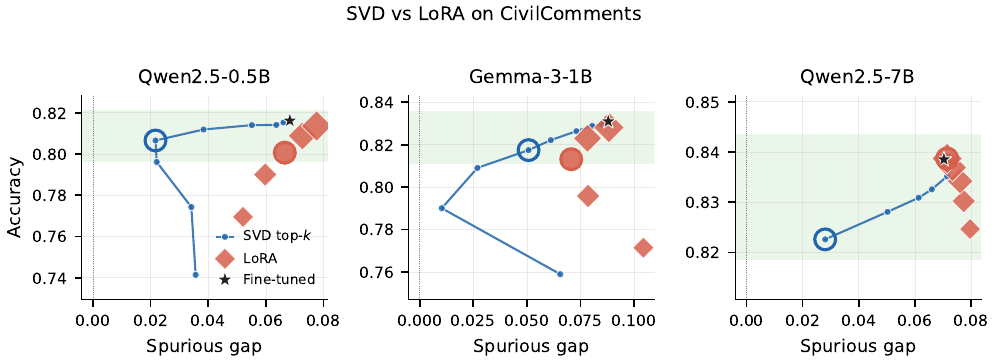}
  \caption{\textbf{Top-$k$ Pareto-dominates LoRA across model scales.} Accuracy versus gap on CivilComments, one panel per model ($0.5$B, $1$B, $7$B). Blue: post-hoc top-$k$ sweep, parametric in $r$. Red diamonds: LoRA rank sweep (marker size $\propto$ rank). Green: no-cost zone (accuracy loss $<\!2$ pp from FT). Rings mark the best in-zone point per method. On all three models the SVD ring sits at lower gap than the LoRA ring at matched accuracy. Per-panel axes.}
  \label{fig:svd-vs-lora}
\vspace{-0.2in}
\end{figure}

The post-hoc-vs.-rank-constrained distinction is what gives top-$k$ its structure. Both methods produce a low-rank effective update but arrive there by different routes. Full SFT optimizes with no rank cap; the update is full-rank with the broadly-distributed spectrum of Fig.~\ref{fig:spectra-app}, dominant patterns at larger singular values and weaker, less consistent patterns in the tail. Post-hoc top-$k$ then drops the tail, where the spectral picture predicts the shortcut response sits. LoRA fixes a rank budget at the start of training, and the optimizer must spend it on whatever minimises training loss, with no incentive to place the shortcut in later-removable directions. The two procedures converge to qualitatively different updates at matched effective rank, and the differences are where the debiasing structure lives. We read this as a consistent account of the Pareto-dominance in Fig.~\ref{fig:svd-vs-lora}, not a proof.

Fig.~\ref{fig:svd-vs-lora} repeats across all models: top-$k$ Pareto-dominates LoRA at matched accuracy on $0.5$B, $1$B, and $7$B. Dominance shrinks (but does not invert) at scale, consistent with larger models packing updates into a tighter top subspace.

\section{Discussion}
\label{sec:discussion}

\paragraph{Takeaway.} A single post-hoc operation, truncating the tail of the SVD of $\Delta W$, reduces the gap on every (model, dataset) cell at $<\!2$ pp accuracy loss, with no labels, retraining, or extra data. We propose this works because the shortcut response sits in the tail of the singular ordering of $\Delta W$: decoupling under truncation is visible across all four natural-shortcut datasets (sharpest on CivilComments), and IMDB-marker realises the predicted boundary, where SFT has only the shortcut to learn and gap and accuracy instead lockstep toward the unbiased base. Bottom-/random-$k$ and matched-rank LoRA rule out generic low-rank approximation and rank-constrained training. The singular basis of $\Delta W$ is a useful coordinate system for asking \emph{what} fine-tuning has learned, not just \emph{how well}.

\paragraph{A working interpretation.} The pattern is consistent with the following picture, which we put forward as a working hypothesis rather than a verified claim. During fine-tuning, dominant and broadly applicable task patterns are absorbed into the top singular components of $\Delta W$, where the optimizer concentrates the largest weight changes. Weaker and less consistent regularities, the kind that produce spurious correlations, such as demographic skew, annotator preferences, or scraping cues, settle into the tail. Post-hoc truncation then removes the tail and with it the shortcut reliance, while leaving the task response largely intact. This is a plausible story for why top-$k$ behaves as it does on the natural-shortcut datasets, and it is consistent with the IMDB-marker boundary, where there is no task signal to land in the top components and so no tail-vs-top separation to exploit. Verifying it requires direct probes of the singular subspaces of $\Delta W$, which we leave to future work.

\paragraph{Limitations.} Our spectral claim is recovered behaviourally with truncation effects in the $(\mathrm{acc}, \Delta)$ plane, not from direct probes of the singular subspaces. While CivilComments is the sharpest decoupling evidence and IMDB-marker realises the predicted boundary, on the other natural NLI/QA datasets we report empirical gap reduction without decomposing how much reflects selective tail removal vs.\ partial reversion toward base. Evaluation is restricted to classification tasks with cleanly defined spurious correlations.

\paragraph{Future work.} Direct probing of the top vs.\ bottom singular subspaces of $\Delta W$ would convert the behavioural claim into a mechanistic one and is the highest-priority follow-up. Per-layer analysis is a natural extension. Applying the diagnostic to complex reasoning, longer generative tasks, and safety-relevant fine-tuning tests the generality of the picture.

\newpage

\section*{Impact Statement}
This work investigates the spectral structure of fine-tuning updates as a tool for mitigating spurious correlations, with fairness as our primary motivating application. By isolating and removing components of the update that encode unwanted shortcuts, our approach offers a principled lens on what fine-tuning actually learns and how undesirable behaviours can be selectively suppressed. We note, however, that the same mechanism is intent-agnostic: if desirable behaviours, such as safety alignment, reasoning capabilities, or task-specific skills, are spectrally separable in a similar way, they could in principle be removed by the same procedure. We view this dual-use possibility as a reason for further study rather than a blocker, since understanding which capabilities are spectrally localized is itself important for building more robust and interpretable models. Beyond fairness, the framework suggests broader benefits: more compact fine-tuning updates and improved generalization, by discarding spectral components that capture dataset-specific noise rather than transferable structure.

\section*{Acknowledgements}
This work was supported by the Modal compute grant.

\bibliography{example_paper}
\bibliographystyle{icml2026}

%%%%%%%%%%%%%%%%%%%%%%%%%%%%%%%%%%%%%%%%%%%%%%%%%%%%%%%%%%%%%%%%%%%%%%%%%%%%%%%
% APPENDIX
%%%%%%%%%%%%%%%%%%%%%%%%%%%%%%%%%%%%%%%%%%%%%%%%%%%%%%%%%%%%%%%%%%%%%%%%%%%%%%%
\newpage

\appendix
\onecolumn

\section{The IMDB-marker control: a predicted boundary of the spectral picture}
\label{app:imdb-control}

\paragraph{Construction.} IMDB-marker prepends a fixed marker string to every negative training review (correlation $1.0$ with the negative label). The fine-tuned model learns to associate the marker with the negative class deterministically. The evaluation set is constructed to be \emph{balanced across $(\text{label}, \text{marker})$ groups}: $50\%$ of positive eval reviews and $50\%$ of negative eval reviews carry the marker. This eval distribution is adversarial to the shortcut by design: a model that has learned ``marker $\Rightarrow$ negative'' from training will misclassify positives that carry the marker and negatives that lack it. As a result the fine-tuned model sits at chance on the balanced evaluation distribution ($\mathrm{acc}_\mathrm{ft} \approx 0.51$) and exhibits a large spurious gap ($\Delta_\mathrm{ft} \approx 0.83$). The base model never saw the marker and is a competent zero-shot sentiment classifier on the same distribution ($\mathrm{acc}_\mathrm{base} \approx 0.88$, $\Delta_\mathrm{base} \approx 0$). The marker is constructed so that it is the only signal fine-tuning can learn, which means the entire fine-tuned update encodes the shortcut.

We emphasise what the FT model has and has not learned. Within the training distribution, where the marker is perfectly correlated with the label, the FT model is correct on every example; it has not become a globally degenerate classifier. What the construction shows is that fine-tuning on this distribution yields a model whose decisions are dominated by a feature that an adversarial eval distribution can break. The chance-level accuracy and large gap are properties of the evaluation, not of the model in isolation. We use this as a controlled boundary case, not as a claim that fine-tuned models in general behave this way; the natural-shortcut datasets in the main body are the realistic regime.

\paragraph{Why this is a prediction, not a separate mechanism.} The spectral-stratification hypothesis (Sec.~\ref{sec:results:mechanism}) says the shortcut response sits in the tail of the singular ordering of $\Delta W$. This presupposes that there \emph{is} a tail to identify, which in turn presupposes that fine-tuning learned more than just the shortcut. If the dataset offers no other signal, that presupposition fails: the entire update encodes the shortcut, and there is no top-vs-tail structure for truncation to exploit. In that case the only path top-$k$ truncation can take is to shrink $\Delta W$ toward zero, which by construction returns the model to base ($\widetilde{W} = W_\mathrm{base}$ at $r=0$). The base model has never seen the marker, so its gap is near zero. The framework therefore predicts a specific signature for this regime: gap and accuracy track each other along the FT-to-base trajectory, rather than decoupling as they would under selective tail removal. IMDB-marker is constructed to test exactly this prediction.

\paragraph{Observed behaviour.} Fig.~\ref{fig:mechanism} in the main body shows the prediction confirmed. CivilComments shows the decoupling signature: gap drops to $\sim\!0.3\,\Delta_\mathrm{ft}$ across the sweet zone while accuracy stays at $\sim\!0.81$. IMDB-marker shows the lockstep signature: accuracy rises smoothly from $0.51$ (FT, near chance) to $0.87$ (close to base $0.88$), and the spurious gap drops from $\Delta_\mathrm{ft}$ to $\sim\!0.04$ at $r{=}5\%$. The two trajectory shapes are visually distinct, and the IMDB shape is the one the framework predicts when the entire update is shortcut.

\paragraph{Why we exclude IMDB-marker from Table~\ref{tab:consistency}.} The table reports bias reduction at the sweet-spot retention $r^*$, defined as the maximum reduction inside the no-cost zone (accuracy loss $<\!2$ pp). On IMDB-marker the FT accuracy is already at chance, so any retention that meaningfully reduces the gap also moves accuracy substantially (along the diagonal toward base), and the no-cost-zone definition becomes ill-defined. Numerically, IMDB-marker shows positive bias reduction across retentions ($\sim\!24\%$ at $r{=}20\%$ on \textsc{Qwen-0.5B}), but reporting it in the same column as the natural-shortcut cells would obscure that the trajectory shape, not just the endpoint magnitude, is qualitatively different. Fig.~\ref{fig:mechanism} reports the trajectory directly.

\section{Additional results}
\label{app:results}

This appendix provides per-task and per-model breakdowns supporting the main-body claims. Conventions: SVD top-$k$ unless otherwise noted; gap reported as a raw value (when absolute scale matters) or normalised by the fine-tuned gap (when comparison across tasks matters); three random seeds aggregated as mean $\pm 1\sigma$.

The main-body trajectory plots (Figs.~\ref{fig:sweet}, \ref{fig:methods}, \ref{fig:svd-vs-lora}) are parametric in retention $r$, projected into the (accuracy loss, gap) plane. The appendix figures here plot each metric directly against $r$, so each curve is a proper function of $r$. Within-curve non-monotonicity reflects the nonlinear dependence of gap and accuracy on which singular components are retained, not seed noise.

\begin{figure*}[p]
  \centering
  \includegraphics[width=0.92\textwidth]{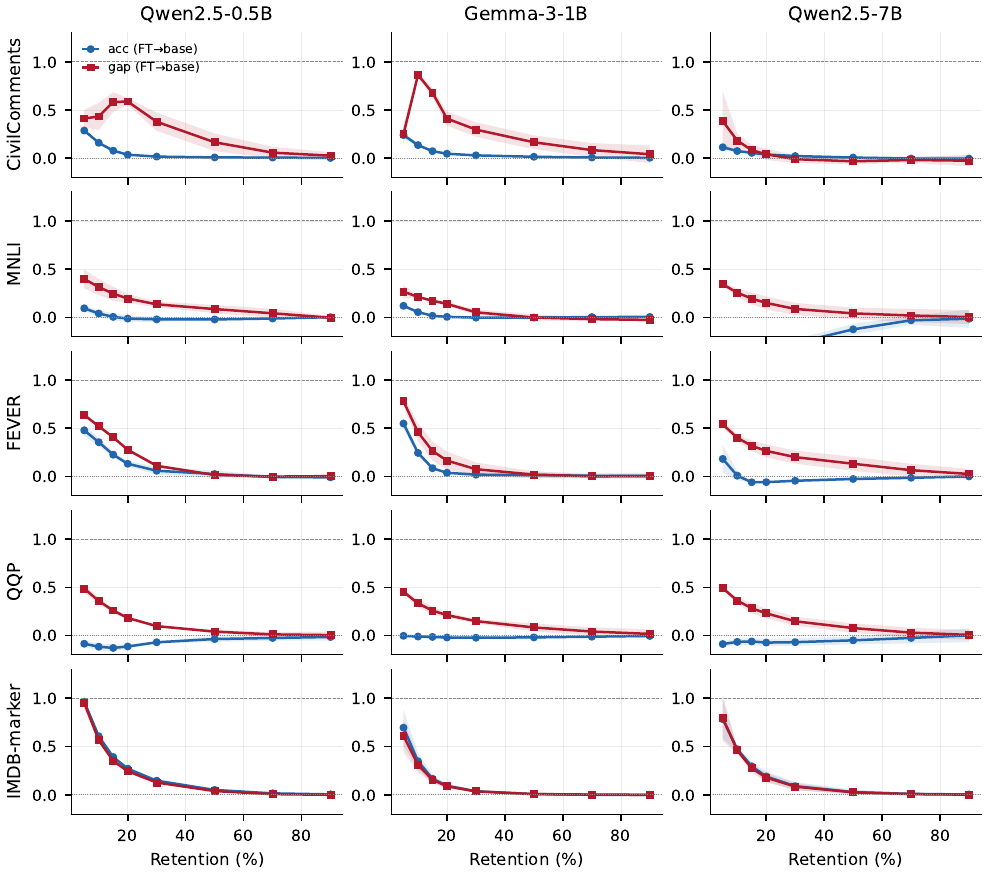}
  \caption{\textbf{Per-(dataset, model) SVD top-$k$ retention sweep.} Both metrics are rescaled to the FT$\to$base interval: $\widetilde{\mathrm{acc}}_r = (\mathrm{acc}_r - \mathrm{acc}_\mathrm{ft})/(\mathrm{acc}_\mathrm{base} - \mathrm{acc}_\mathrm{ft})$ and $\widetilde{\Delta}_r = (\Delta_r - \Delta_\mathrm{ft})/(\Delta_\mathrm{base} - \Delta_\mathrm{ft})$, so $0$ corresponds to FT and $1$ to base on each axis. Bands: $\pm 1\sigma$ over three seeds. Top four rows (CivilComments, MNLI, FEVER, QQP): \emph{decoupling}, with blue (acc) staying near $0$ while red (gap) rises toward $1$, i.e.\ accuracy is preserved at the FT level while the gap is pulled toward the unbiased base. Bottom row (IMDB-marker, boundary case): \emph{lockstep}, with both curves rising together from $0$ to $1$, the FT-to-base trajectory predicted when $\Delta W$ encodes the shortcut alone.}
  \label{fig:sweeps-app}
\end{figure*}

\paragraph{Per-(model, dataset) retention sweep.} Fig.~\ref{fig:sweeps-app} reports the spurious gap and overall accuracy as functions of retention for every (model, dataset) cell. Both metrics are rescaled to the FT$\to$base interval ($0$ = FT, $1$ = base) so the two metrics live on the same axis and the diagnostic shapes from Sec.~\ref{sec:results:mechanism} are directly visible. Bands are $\pm 1\sigma$ over three seeds. On the natural-shortcut datasets (CivilComments, MNLI, FEVER, QQP), accuracy (blue) stays near $0$ across retentions while the gap (red) rises toward $1$: the gap is pulled toward the unbiased base while accuracy is preserved at the FT level, the decoupling signature predicted by spectral stratification. IMDB-marker (bottom row) shows the boundary signature instead: with the marker the only signal SFT can learn, no top-vs-tail structure exists, and both curves rise together from $0$ to $1$ along the FT-to-base trajectory. The two trajectory shapes (decoupling on natural-shortcut datasets, lockstep on IMDB-marker) are the appendix-scale view of Fig.~\ref{fig:mechanism}.

\begin{figure*}[t]
  \centering
  \includegraphics[width=\textwidth]{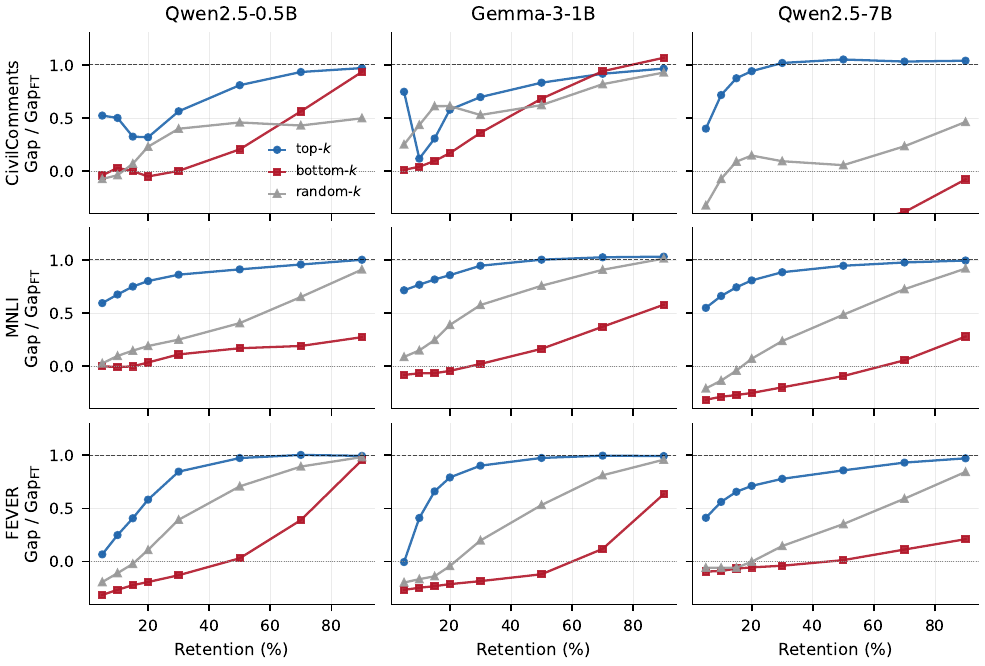}
  \caption{\textbf{Top-$k$ vs.\ bottom-$k$ vs.\ random-$k$} on three representative datasets crossed with three models. Y-axis: normalised gap $\Delta_r/|\Delta_\mathrm{ft}|$; dashed line at $1.0$ marks the fine-tuned reference. Top-$k$ approaches $0$ smoothly; bottom-$k$ and random-$k$ either stay near $1.0$ until accuracy collapses, or overshoot below $0$ as the model reverts toward an unbiased base.}
  \label{fig:methods-app}
\end{figure*}

\paragraph{Method comparison on representative datasets.} Fig.~\ref{fig:methods-app} replicates the method comparison of Sec.~\ref{sec:results:methods} on three datasets across all three models, plotting normalised gap ($\Delta_r / |\Delta_\mathrm{ft}|$): $1.0$ means no debiasing, $0$ means fully debiased. Top-$k$ reaches near-zero normalised gap at low retention while preserving accuracy. Bottom-$k$ and random-$k$ overshoot below zero at sufficiently small $k$, a signature of reversion toward base rather than selective shortcut removal.

\begin{figure*}[t]
  \centering
  \includegraphics[width=\textwidth]{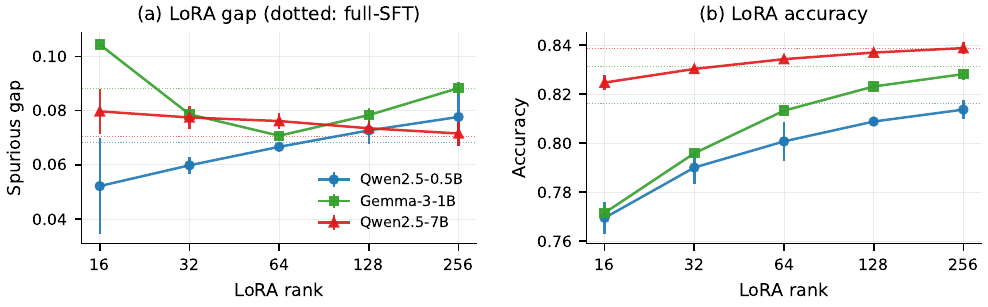}
  \caption{\textbf{LoRA rank sweep on CivilComments.} Dotted lines: per-model full-SFT reference. Low-rank LoRA points can fall below the SFT gap reference, but only because they also fall below the SFT \emph{accuracy} reference (right panel): the gap is reduced by underfitting, not by selectively removing the shortcut. The matched-accuracy comparison in Fig.~\ref{fig:svd-vs-lora} is the apples-to-apples view. Accuracy rises with rank toward the SFT level.}
  \label{fig:lora-app}
\end{figure*}

\paragraph{LoRA rank sweep across models.} Fig.~\ref{fig:lora-app} shows the LoRA rank sweep on CivilComments for all three models. The comparison that matters is at \emph{matched accuracy}, which Fig.~\ref{fig:svd-vs-lora} reports directly: top-$k$ Pareto-dominates LoRA on all three models. The per-rank view here is the supporting decomposition. At small ranks (e.g.\ $r{=}16$ on \textsc{Qwen-0.5B}), LoRA can show a gap below the full-SFT reference, but only because it has not yet recovered full-SFT accuracy; the low gap is bought by underfitting the task, not by selectively removing the shortcut. As rank rises, accuracy approaches the full-SFT level and the gap rises toward (or above) the full-SFT reference. The takeaway is that LoRA does not reach a regime where it simultaneously matches full-SFT accuracy and reduces the spurious gap, which is exactly the regime post-hoc top-$k$ truncation occupies. This is consistent with the reading in Sec.~\ref{sec:results:methods}: rank-constrained training packs task and shortcut into a shared low-rank subspace, while unconstrained fine-tuning yields a full-rank update in which post-hoc truncation can selectively drop the tail.

\begin{figure*}[p]
  \centering
  \includegraphics[width=0.92\textwidth]{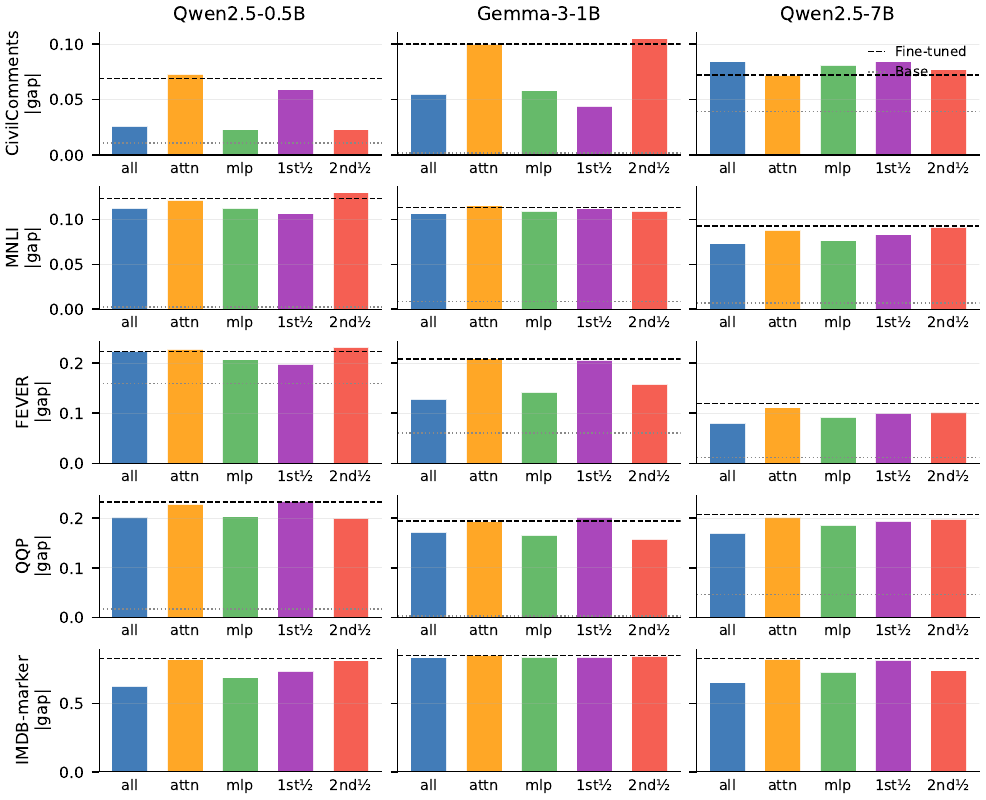}
  \caption{\textbf{Per-layer subset compression at $r{=}20\%$}, real data, per-cell. Bars show $|\Delta|$ when only the indicated layer subset is truncated; remaining layers retain the full-rank update. Dashed: fine-tuned $|\Delta|$; dotted: base $|\Delta|$. Some cells show clean MLP- or second-half-localised reduction (e.g.\ CivilComments on \textsc{Qwen-0.5B}); others show the relevant directions spread across the network (e.g.\ CivilComments on \textsc{Qwen-7B}). We do not claim a universal localisation.}
  \label{fig:perlayer-app}
\end{figure*}

\paragraph{Per-layer subset compression.} Fig.~\ref{fig:perlayer-app} restricts truncation at $r{=}20\%$ to a subset of layers (attention only, MLP only, first half, second half), keeping the full-rank update elsewhere. We use this as an exploratory probe of where in the network the shortcut-related component of $\Delta W$ lives. The pattern is heterogeneous across (model, dataset) cells. On some cells the MLP-only or second-half subsets recover much of the bias reduction of full truncation (e.g.\ CivilComments on \textsc{Qwen-0.5B}), suggesting the relevant directions concentrate in those layers. On other cells (e.g.\ CivilComments on \textsc{Qwen-7B}, MNLI on \textsc{Qwen-0.5B}) no single subset recovers a substantial fraction of the full-truncation reduction, indicating the relevant directions are distributed across the network. We do not claim a universal layer-localisation result; we report the experiment because the heterogeneity is itself informative about how fine-tuning organises the update. On IMDB-marker the subsets behave heterogeneously in a way consistent with the predicted boundary regime (App.~\ref{app:imdb-control}): truncation contributes to the FT-to-base trajectory wherever it is applied, since by construction there is no top-vs-tail structure to exploit on this dataset.

\begin{figure*}[t]
  \centering
  \includegraphics[width=\textwidth]{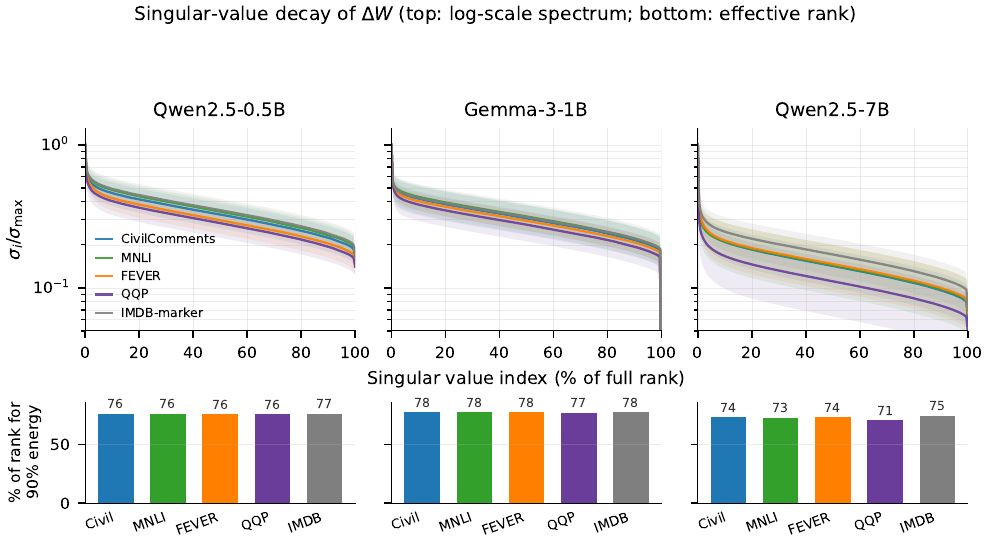}
  \caption{\textbf{Real singular-value decay of $\Delta W$} for four representative MLP layers (mean $\pm 1\sigma$ shaded band). Top row: $\sigma_i / \sigma_{\max}$ on a log $y$-axis, all five datasets overlaid per model. Bottom row: percentage of singular components needed to capture $90\%$ of the spectral energy. Spectra are similar across datasets within a model and are not sharply concentrated ($90\%$ of energy needs $\sim\!73\!-\!78\%$ of components). $\Delta W$ is therefore not approximately low-rank, which rules out the naive ``shortcut has small spectral energy'' reading and motivates the directional / ordering reading discussed above.}
  \label{fig:spectra-app}
\end{figure*}

\paragraph{Singular-value decay of $\Delta W$.} Fig.~\ref{fig:spectra-app} shows the real singular-value spectrum of $\Delta W$ averaged across four representative MLP layers, with all five datasets overlaid per model. Two observations follow. First, the spectra are nearly indistinguishable across datasets within a given model: the difference between cells where the spectral picture applies cleanly (CivilComments) and the predicted boundary regime where it is forced to break down (IMDB-marker) is \emph{not} visible in the raw spectrum. Second, $\Delta W$ is \emph{not} approximately low-rank: $90\%$ of the spectral energy lives in roughly $73\!-\!78\%$ of the singular components, so the top few singular values do not dominate the variance.

This second observation reinforces the framing in Sec.~\ref{sec:results:mechanism}. The naive reading (``$\Delta W$ is low-rank; task signal lives in the top singular values; the shortcut lives in a tiny tail; drop the tail'') is not supported by the spectrum. The supported reading is the directional / behavioural one. Writing $\Delta W = \sum_i \sigma_i u_i v_i^\top$, top-$k$ truncation preserves the model's response to inputs whose projection onto $v_1, \ldots, v_k$ is large and removes the response to inputs whose projection lies in the bottom subspace $v_{k+1}, \ldots, v_n$. The empirical finding that truncation preserves task accuracy while reducing the spurious gap therefore implies that task-relevant input directions project predominantly onto the top right-singular vectors of $\Delta W$, and shortcut-related input directions project predominantly onto the bottom. This is a claim about \emph{ordering} in the singular basis, recovered from the effect of truncation, not a claim about energy concentration. The spectrum can be broadly distributed (as it is in real data) and the directional property can still hold; the claim is then ``the shortcut response sits in the tail of the singular ordering'', not ``the shortcut response carries little spectral energy''. We do not claim the property is visible in the raw spectrum; Fig.~\ref{fig:spectra-app} confirms that it is not. The behavioural sweeps (Figs.~\ref{fig:sweet}, \ref{fig:methods}) are the direct evidence.

\end{document}